\RequirePackage{fix-cm}
\documentclass[twocolumn]{svjour3}          % twocolumn
\smartqed  % flush right qed marks, e.g. at end of proof
\usepackage{graphicx}
\usepackage{interval}
\usepackage{multirow}
\usepackage{url}

\journalname{Machine Vision and Applications}

\begin{document}

\title{Detection of 3D Bounding Boxes of Vehicles Using Perspective Transformation for Accurate Speed Measurement}


\author{Viktor~Kocur         \and
        Milan~Ft\'{a}\v{c}nik %etc.
}


\institute{V. Kocur \\ \email{viktor.kocur@fmph.uniba.sk} \\ \\
Department of Applied Informatics, Faculty of Mathemathics, Physics and Informatics of Comenius University in Bratislava, Mlynská Dolina, 841 01, Bratislava, Slovakia}

\date{Received: date / Accepted: date}
%\date{ }

\maketitle

\begin{abstract}
Detection and tracking of vehicles captured by traffic surveillance cameras is a key component of intelligent transportation systems. We present an improved version of our algorithm for detection of 3D bounding boxes of vehicles, their tracking and subsequent speed estimation. Our algorithm utilizes the known geometry of vanishing points in the surveilled scene to construct a perspective transformation. The transformation enables an intuitive simplification of the problem of detecting 3D bounding boxes to detection of 2D bounding boxes with one additional parameter using a standard 2D object detector. Main contribution of this paper is an improved construction of the perspective transformation which is more robust and fully automatic and an extended experimental evaluation of speed estimation. We test our algorithm on the speed estimation task of the BrnoCompSpeed dataset. We evaluate our approach with different configurations to gauge the relationship between accuracy and computational costs and benefits of 3D bounding box detection over 2D detection. All of the tested configurations run in real-time and are fully automatic. Compared to other published state-of-the-art fully automatic results our algorithm reduces the mean absolute speed measurement error by 32\% (1.10 km/h to 0.75 km/h) and the absolute median error by 40\% (0.97 km/h to 0.58 km/h).

%for a semi-automatic method and 30\% (1.10 km/h to 0.77 km/h) and 40\% (0.97 km/h to 0.58 km/h) for a fully automatic method.


\end{abstract}

\section{Introduction}
Recent development in commercially available cameras has increased the quality of recorded images and decreased the costs required to install cameras in traffic surveillance scenarios. Automatic traffic surveillance aims to provide information about the surveilled vehicles such as their speed, type and dimensions and as such is an important aspect of intelligent transportation system design.

Automatic traffic surveillance system requires an accurate way of detecting the vehicles in the image and an accurate calibration of the recording equipment. 

Standard procedures of camera calibration require a calibration pattern or measurement of distances on the road plane. Dubsk\'{a} et al. \cite{dubska2014} proposed a fully automated camera calibration for the traffic surveillance scenario. We use an improved version \cite{sochor2017} of this method to obtain the camera calibration and focus on the accuracy of vehicle detection.

Object detection is one of the fundamental tasks of computer vision. Recent deep learning techniques have successfully been applied to this task. Deep convolutional neural networks are used to extract features from images and a supplementary structure utilizes these features to detect objects. We opt to use the object detector RetinaNet \cite{RetinaNet} as a base framework for object detection as it offers good tradeoff between accuracy and low inference times. RetinaNet uses a structure of so-called anchor boxes for object detection and our method could therefore utilize any other widely used object detection framework based on anchor boxes \cite{SSD,redmon,FasterRCNN}. With minor modifications our method could also utilize emerging object detection frameworks based on keypoint detection \cite{CornerNet,ObjectsAsPoints}.

In this paper we extend our previous work \cite{CVWW2019} where we proposed a perspective image transformation which utilizes the geometry of vanishing points in a standard traffic surveillance scenario. The perspective transformation enables us to rectify the image which has two significant effects. The first effect is that this aids the object detection accuracy. The second one is that this enables an intuitive parametrization of the 3D bounding box of a vehicle as a 2D bounding box with one additional parameter. Our method has surpassed the existing state-of-the-art approaches in terms of speed measurement accuracy while being computationally cheaper. The method was mostly automatic, but the construction of the perspective transformation was not robust enough resulting in a need for manual adjustments for some camera angles. Now we propose a new approach which remedies this problem and enables two different transformations to be constructed for a single traffic scene. We also provide an extended study of performance of our method with different configurations to gauge their effects on the speed measurements accuracy and computational costs to offer various options for different computational constraints. We also show that the improved transformation brings improvements in speed measurement accuracy.

\section{Related Work}

Measuring speeds of vehicles captured by a monocular camera requires their detection and subsequent tracking followed by measurement of the distance they passed utilizing camera calibration. Connecting these subtasks into a single pipeline is usually trivial so in the last subsection we focus on the available means of evaluating the accuracy of the whole pipeline.

\subsection{Object Detection}

Recent advent of convolutional neural networks had a significant impact on the task of object detection. Two stage approaches such as Faster R-CNN \cite{FasterRCNN} use a convolutional neural network to generate proposals for objects in image. In the second stage the network determines which of these proposed regions contain objects and regress the boundaries of their bounding boxes. 

Single stage approaches \cite{RetinaNet,SSD,redmon} work by using a structure of anchor boxes as the output of the network. Each anchor box represents a possible bounding box. Each anchor box has a classification output to determine which object, if any, is in the anchor box and a regression output to align the bounding box to the object. In this approach one object can be covered by multiple anchor boxes so a technique such as non-maximum suppression must be used to leave only one bounding box per object. 

Current state of the art approaches forego the use of anchor boxes completely and rely on detecting keypoints in the image via heatmaps on the output of the network. CornerNet \cite{CornerNet} detects the two opposite corners of a bounding box and pairs them using an embedding. CenterNet \cite{CenterNet} detects the center of the object and uses regression to determine the dimensions of the object.

\subsection{Vehicle Detection}

%Vehicles can be detected in various forms for purposes of traffic surveillance. Most cameras employed in traffic surveillance are static. Many vehicle detection approaches \cite{dubska2014,slotcars,maduro} therefore utilize background subtraction methods to detect the vehicles. These methods can fail with quickly changing lighting conditions or small camera movements and may result in single detections containing more than one vehicle due to occlusion. Luviz\'{o}n et al. use motion detector \cite{frommotion} to detect license plates. Daley et al. \cite{dailey} find edges in the inter-frame differences. Pham and Lee \cite{windsheld} propose a method based on finding the windshields of vehicles. Zhou et al. \cite{dave} propose two deep neural networks to propose and verify the detections. 

In our work we focus on detecting vehicles via their 3D bounding boxes as this approach has been shown to be also beneficial for subsequent tasks such as fine-grained vehicle classification \cite{boxcars,perspectivenet} and re-identification \cite{vehiclereid}. In the evaluation section of this paper we also show that detecting 3D bounding boxes as opposed to 2D bounding boxes is beneficial to speed measurement accuracy. 

Background subtraction is a common method of detecting vehicles as the traffic surveillance cameras are static. Corral-Soto and Elder \cite{slotcars} fit a mixture model for the distribution of vehicle dimensions on labeled data. The model is used together with the known geometry of the scene to estimate the vehicle configuration for blobs of vehicles obtained via background subtraction. Similarly, Dubsk\'{a} et al. use background subtraction to obtain masks of vehicles. 3D bounding boxes aligned with the vanishing points of the scene are then constructed tangent to these masks. The order of construction of the edges of the bounding box is important and the process may not be stable. This approach has been slightly improved \cite{sochor2017} by using Faster R-CNN object detector \cite{FasterRCNN} before the background subtraction to determine which blobs are cars. Approaches relying on background subtraction can be sensitive to changing light conditions or vibrations of traffic cameras and may thus not be suitable for some traffic surveillance scenarios.

Zeng et al. \cite{perspectivenet} use a combination of two networks to determine the 3D bounding boxes of vehicles, which are subsequently used to aid in the task of fine-grained vehicle classification. The first network is based on RetinaNet object detector \cite{RetinaNet}. The second network is given the position of 2D bounding boxes obtained by the first network to perform a ROIAlign operation \cite{MaskRCNN} on a feature map from a separate ResNet network \cite{resnet}. This second network then outputs the positions of the vertices of the 3D bounding boxes and is trained as standard regression task with a regularization term which ensures that the bounding box conforms to perspective geometry. The obtained geometry of the vehicle is then used to extract features for a fusion network which is trained on the task of fine-grained vehicle classification. The whole system is trained on the BoxCars116k \cite{boxcars} dataset, which contains over 116 thousand images each with one vehicle with annotations containing its 2D and 3D bounding box as well as its make and model. Since the dataset contains 3D bounding box annotations we also utilize it for training.

Multiple approaches for detecting 3D bounding boxes have been published and evaluated on the KITTI dataset \cite{KITTI}. This dataset contains videos from the ego-centric view from a vehicle driving in various urban environments. The videos are annotated with 3D bounding boxes of relevant objects such as cars, cyclists and pedestrians. 

Many published approaches rely on modified 2D object detectors. The authors of CenterNet object detector published \cite{CenterNet} an evaluation of a slightly modified version of their detector on the KITTI dataset. Mousavian et al. \cite{mousavian} use a 2D bounding box and regress orientation and dimensions of vehicles separately and combine them with geometry constraints to obtain a final 3D bounding box. MonoDIS \cite{simonelli} works by adding a 3D detection head on top of a RetinaNet object detector \cite{RetinaNet}. The detection head is trained to regress 10 parameters of the 3D bounding box in a special regime where in each step some parameters are fixed to the ground truth for loss computation. Kim and Kum \cite{kim} propose to use perspective transformation on the image to create a rectified birds eye view of the road plane and find the bounding boxes of vehicles in the transformed image. GS3D \cite{gs3d} uses the Faster R-CNN \cite{FasterRCNN} object detector to find a 2D bounding box of a vehicle. Based on statistics of the test set and some geometrical observations of the common self-driving scenario a rough estimate of the 3D bounding box is generated. Information from the rough 3D bounding box is then used to guide further feature extraction with the use of perspective transformation. The extracted features are then used to refine the 3D bounding box. SMOKE \cite{smoke} uses a structure similar to CenterNet to detect a keypoint in the projected center of the 3D bounding box and regress the parameters of the 3D bounding box. The network is trained end-to-end using a disentanglement loss similar to MonoDIS. MonoPair \cite{monopair} also utilizes a network inspired by CenterNet to detect the 2D and 3D bounding boxes of vehicles. Additionally, the network also detects keypoints which represent the middlepoints between pairs of neighboring vehicles along with regression of the 3D distance of the two vehicles. The network is then trained end-to-end in an uncertainty-aware manner \cite{uncertain} with the loss incorporating a pairwise spatial constraint imposed on the detected vehicles and middlepoints.

\begin{figure*}[t]
   \centering
   \includegraphics[width=\textwidth]{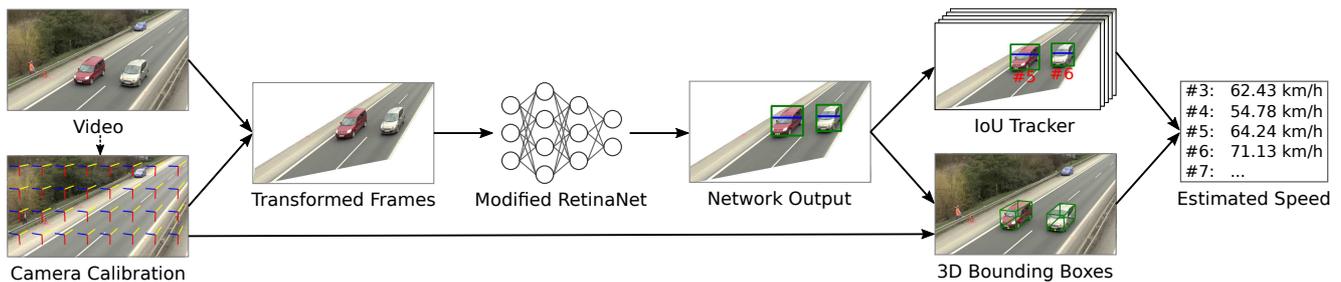}
   \caption{The diagram of our speed estimation pipeline.}
   \label{fig:diagram}
\end{figure*}

The traffic surveillance scenario is significantly different from the autonomous vehicle scenario of the KITTI dataset due to lack of constancy of scene and thus its geometry as well as a different vantage point of the camera. It is therefore not possible to compare approaches for these two tasks directly. Our method shares similarities to the presented works by using a 2D object detector, while exploiting additional constraints that can be assumed in a geometry of a standard traffic surveillance scenario.

\subsection{Camera Calibration}

In the context of traffic surveillance, camera calibration is necessary to enable measurement of real world distances in the surveilled scene. A review of available methods has been presented by Sochor et al. \cite{brnocompspeed}. The review found that most published methods are not automatic and require human input such as drawing a calibration pattern on the road \cite{pattern}, using positions of line markings on the road \cite{cathey,lan,maduro} or some other measured distance related to the scene \cite{schoepflin,you}.

Ideally, camera calibration can be performed automatically and accurately. Filipiak et al. \cite{filipiak} proposed an automatic approach based on an evolutionary algorithm, though the approach was validated only on footage zoomed in to obtain clear image of license plates, which is unsuitable for traffic surveillance on multi-lane roads.  

A fully automatic method has been proposed by Dubsk\'{a} et al. \cite{dubska2014}. The camera is calibrated by finding the three orthogonal vanishing points related to the road plane. The first vanishing point corresponds to the movement of the vehicles. Relevant keypoints are detected and tracked using the KLT tracker. The tracked lines of motion are then transformed into a so-called diamond space based on parallel coordinates in a fashion similar to the Hough transform. Edges of vehicles which are perpendicular to their movement are used in the same way to determine the position of the second vanishing point. Under the assumption that the principal component is in the center of the image the focal length of the camera is then calculated and subsequently the last vanishing point is determined to be perpendicular to the first two using vector product in homogeneous coordinates. To enable measurements of distances in the road plane a scale factor needs to be determined. The dimensions of the detected vehicles are recorded and their mean is calculated. The mean is compared to statistical data based on typical composition of traffic in the country to obtain the scale. This method has been further improved by Sochor et al. \cite{sochor2017} by fitting a 3D model of a known common vehicle to its detection in the footage. The detection of the second vanishing point is also improved by using edgelets instead of edges. We opt to use this improved fully-automatic calibration method in our pipeline.

\subsection{Object Tracking}

To allow for vehicle counting and speed measurement, the vehicles have to be tracked from frame to frame. Since object detection may sometimes fail a robust tracker is necessary. Kalman filter \cite{Kalman} has been a reliable tool to tackle the task of object tracking in many domains. Bochinksi et al. \cite{IOU} have shown that a simple IoU tracker can outperform more complicated trackers when the objects are detected reliably. Based on this we choose to opt for a similar tracking strategy.

\subsection{Visual Traffic Speed Measurement Datasets}

A review \cite{brnocompspeed} of existing traffic camera calibration methods, vehicle speed measurement methods and evaluation datasets found that many of the published results are evaluated on small datasets with ground truth known for only few of the surveilled vehicles. Additionally, most of the datasets used in published literature were not publicly available. The authors of the review offer their own dataset called BrnoCompSpeed containing 21 one hour long videos which collectively contain 20 thousand vehicles with known ground truth speeds obtained via laser gates. The authors also provide an evaluation script for this datset. We choose to perform the main evaluation of our method on this dataset.

Luviz\'{o}n et al. \cite{luvizon} have published a dataset containing 5 hours of footage from a single intersection. The dataset includes ground truth speeds of vehicles measured by inductive loops installed in the road as well as annotated positions of the license plates of vehicles. The authors provide their own pipeline for speed estimation which is based on detection of license plates. The license plates are detected by generating candidate regions around horizontal edges of moving vehicles. These regions are then validated using a T-HOG \cite{thog} descriptor and an SVM classifier. After the license plate is detected a few keypoints located within it are tracked using a pyramidal KLT tracker \cite{pKLT}. A manually obtained homography matrix is used to determine the real world coordinates of the tracked keypoints and thus also the speed of the vehicles.

\section{Proposed Algorithm}

The goal of our algorithm is to detect 3D bounding boxes of vehicles recorded with a monocular camera installed above the road plane, track the vehicles and evaluate their speed. The algorithm consists of several stages and it requires the camera to be calibrated as per \cite{sochor2017}. The algorithm is an improvement of our previous work \cite{CVWW2019}.

We will first provide an overview of the whole algorithm and then describe each part in greater detail. The diagram of the whole pipeline can be seen in Figure \ref{fig:diagram}. At first, a perspective transformation is constructed using the positions of the vanishing points of the traffic scene which are known thanks to the calibration. This transformation is applied to every frame of the recording. The transformed frames are used as an input to a RetinaNet object detector which detects vehicles and their 2D bounding boxes with one additional parameter. The output of the object detection network is used for tracking and 3D bounding box construction. Tracking is performed by comparison of the 2D bounding boxes in successive frames using a simple algorithm based on the IoU metric. The 3D bounding boxes are constructed in the transformed frames using the 2D bounding boxes with the additional parameter. Inverse perspective transformation is then used to transform the 3D bounding boxes onto the original scene. The center of the bottom frontal edge of every 3D bounding box is used to provide pixel position of a vehicle for each frame. The calibration is then used to project the pixels onto the road plane and thus enable measurement of the distance traveled between frames by one vehicle. These interframe distances are consequently used to measure the speed of the vehicle over the whole track.

\subsection{Image Transformation}
\begin{figure*}[]
   \centering
   \includegraphics[width=\textwidth]{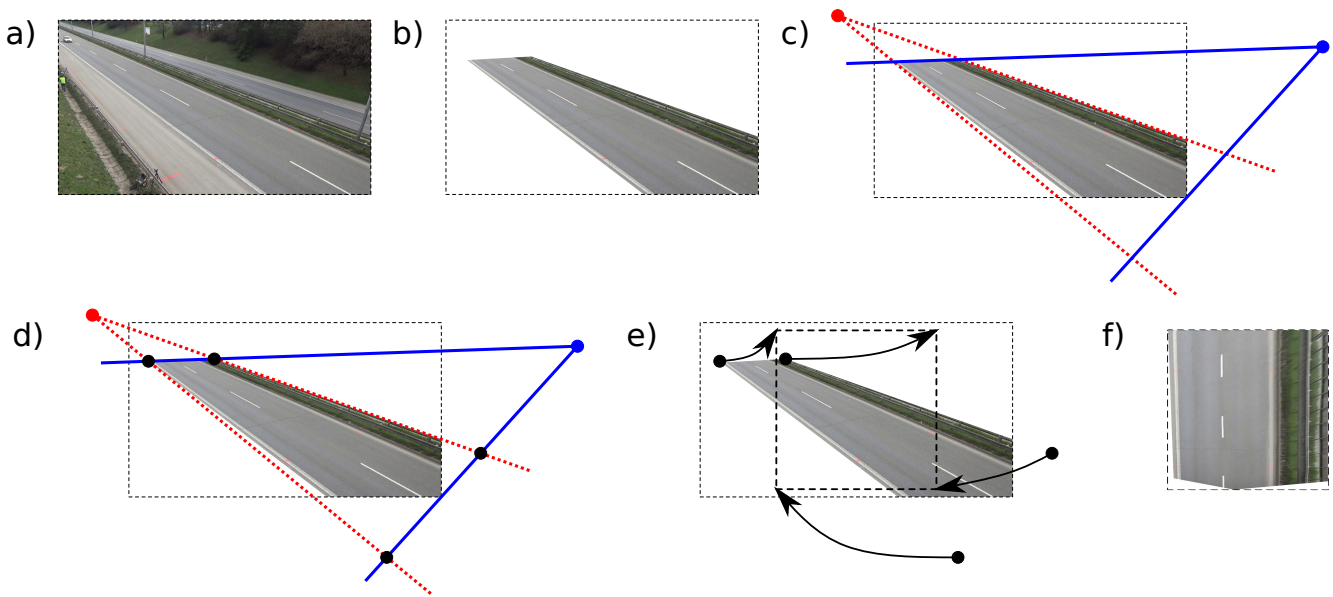}
   \caption{The process of the construction of the perspective transformation for the pair \textit{VP1-VP2}. \textbf{a)} The original traffic surveillance scene. \textbf{b)} The mask of the desired road segment is applied. \textbf{c)} For both \textit{VP1} (dotted red) and \textit{VP2} (solid blue) lines which originate in the vanishing point and are tangent to the mask from both sides are constructed. \textbf{d)} The four intersections of these lines are found. \textbf{e)} The four points are paired with the four corners of the rectangle with the desired dimensions of the transformed image. \textbf{f)} When the transformation is applied the lines corresponding to each of the two vanishing points are parallel to the axes. If the total blank (white) area in the transformed image is more than 20\% of the pixels in the transformed image then the mask is cropped by few pixels from the bottom and the process starts again from step b).}
   \label{fig:transformation}
\end{figure*}

To construct the image transformation we require the camera to be calibrated as described in \cite{sochor2017}. This calibration method has very few limitations regarding the camera position. The camera has to be positioned above the road plane and the observed road segment has to be straight. The main parameters obtained by the calibration are the positions of the two relevant vanishing points in the image. Assuming that the principal point is in the center of the image, the position of the third vanishing point as well as focal length of the camera can be calculated. This enables us to project any point in the image onto the road plane. To enable measurements of distances on the road plane one additional parameter, denoted as scale, is determined during calibration.

The first detected vanishing point (denoted further as \textit{VP1}) corresponds to the lines on the road plane which are parallel to the direction of the moving vehicles. The second detected vanishing point (\textit{VP2}) corresponds to the lines which lie on the road plane but are perpendicular to the the direction of the moving vehicles. The third vanishing point (\textit{VP3}) corresponds to the lines which are perpendicular to the road plane.

The goal of the transformation is to create a new image in which lines corresponding to one of the vanishing points are parallel to one of the image axes and lines corresponding to another vanishing point are parallel to the other image axis. In the transformed image the two vanishing points will thus be ideal points. We also require that the lines corresponding to the last vanishing point remain lines in the transformed image. In order to fulfill these conditions we use the perspective transformation. Since the orientation of the vehicles is closely related to the positions of the vanishing points the vehicles will be rectified in the transformed image.

In our previous work \cite{CVWW2019} we proposed an algorithm which was able to construct such transformation for the pair of vanishing points \textit{VP2} and \textit{VP3}, however we observed that for some camera positions the results were much worse than for the rest. This was caused by an inadequate perspective transformation which resulted in a very small and distorted part of the transformed image to be relevant for detection. At that time we remedied this by significant manual adjustments, which were not automated and therefore undesirable. Furthermore the previous approach failed completely to construct a reasonable transformation for the pair of vanishing points \textit{VP1} and \textit{VP2}.

Now we propose to remedy this problem in an automated fashion by setting a condition that the transformed image should contain as much relevant information as possible. To satisfy this we propose two following adjustments. Firstly, we use a mask of the surveilled traffic lanes for the construction of the transformation instead of the whole image. For evaluation we use the BrnoCompSpeed dataset \cite{brnocompspeed} in which the masks are already provided. In other cases the masks can be obtained automatically by utilizing optical flow \cite{opticalflow}. Secondly, we heuristically set a limit that no more than 20\% of the pixels in the transformed image should correspond to pixels which lie outside of the mask in the original image. In the evaluation section we show that this approach not only makes the algorithm fully automatic, but also leads to better accuracy for the speed estimation task.

The construction algorithm of the transformation has the following steps:
\begin{enumerate}
\item Out of the three vanishing points choose either the pair \textit{VP1-VP2} or \textit{VP2-VP3}.
\item For each of the selected vanishing points construct two lines which originate in the vanishing point and are tangent to the mask. Thus creating four lines.
\item Find the four intersections of the lines, for the pairs of lines which originate in different vanisihng points.
\item Pair each of the four intersection points with a corner of a rectangle with the desired dimensions of the transformed image in the way that preserves the vertical direction of the vehicle movement (e.g. vehicles traveling from top-left to bottom-right will be traveling from top to bottom in the transformed image).
\item Use the four pairs of points to obtain the perspective transformation.
\item Apply the transformation on image of the mask. If the area of the transformed mask is less than 80\% of the transformed image then the original mask is cropped from the bottom by one pixel (if not possible terminate with failure) and the process is repeated from step 2). Otherwise output the transformation.
\end{enumerate}

The algorithm is visualised in Fig. \ref{fig:transformation}. Note that this algorithm may terminate with failure. This is usually the case when the line connecting the two vanishing points intersects the mask. This is problematic since after the transformation the line would have to be transformed to a line that would be perpendicular to itself. This is theoretically possible by transforming the line to infinity, but such a transformation would not be useful. The only way to resolve this is by reducing the mask so that it does not contain the line connecting the vanishing points. In that case even if the algorithm would output a transformation only a small part of the relevant road segment would be visible in the transformed image and would thus be unusable for traffic surveillance. In the BrnoCompSpeed dataset \cite{brnocompspeed} this does not occur. Since the calibration method requires that the three vanishing points are orthogonal it is safe to assume that the transformation would work well for at least one of the pairs.

The method can theoretically be extended to include the transformation for the pair \textit{VP1-VP3}, but on the BrnoCompSpeed dataset this results in failure of the construction algorithm for multiple videos due to the fact that the line between the two vanishing points usually intersects the scene. Therefore, we dismiss this approach. Removing this pair also has the benefit of simplifying the parametrization of the 3D bounding box presented in the following subsection.

\subsection{Parametrization of the 3D Bounding Boxes}

\begin{figure*}[t]
   \centering
   \includegraphics[width=\textwidth]{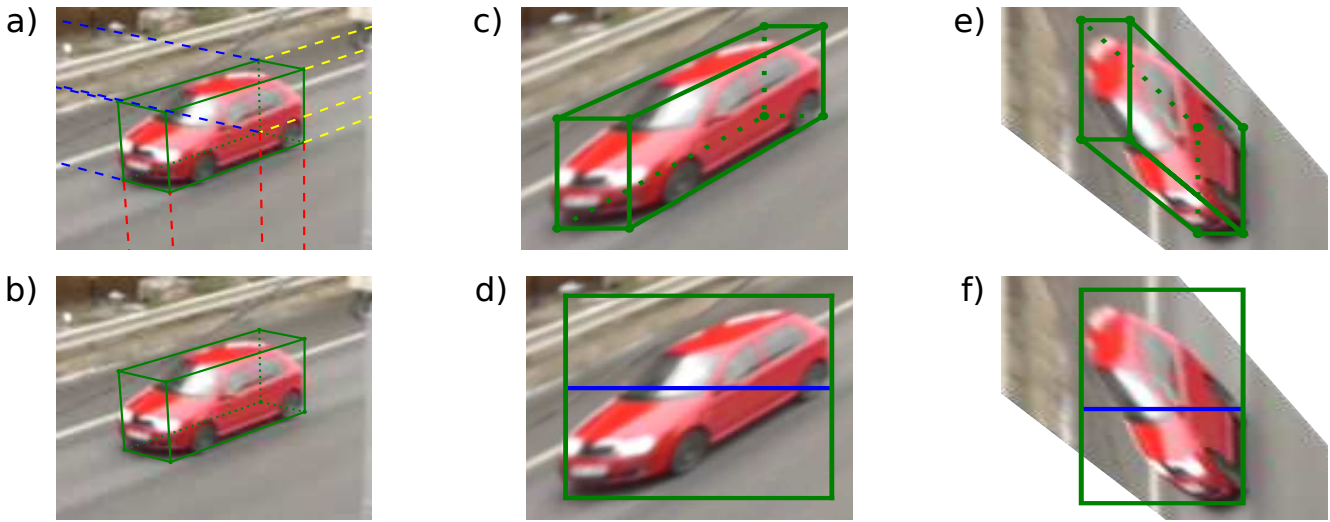}
   \caption{The process of constructing 2D bounding box with the $c_c$ parameter from a 3D bounding box using the transformation for both pairs for a vehicle from the BoxCars dataset \cite{boxcars}. \textbf{a)} 3D bounding box (green) which is aligned with \textit{VP1} (yellow), \textit{VP2} (blue) and \textit{VP3} (red). \textbf{b)} 3D bounding box. \textbf{c)} 3D bouding box after the perspective transform for the pair \textit{VP2-VP3} is applied. \textbf{d)} The parametrization of the 3D bouding box for the pair \textit{VP2-VP3} as a 2D bouding box (green) with the parameter $c_c$ is determined as the ratio of the distance from top of the 2D bounding box to the top-front edge of the transformed 3D bounding box (blue) and the height of the 2D bouding box. \textbf{e)} 3D bouding box after the perspective transform for the pair \textit{VP1-VP2} is applied. \textbf{f)} The parametrization of the 3D bouding box for the pair \textit{VP1-VP2}.}
   \label{fig:bb_construction}
\end{figure*}

We aim to detect 3D bounding boxes aligned with the vanishing points. After performing the image transformation from the previous section, 8 of the 12 edges of the bounding box are aligned with the image axes. This enables us to describe the 3D bounding box as a 2D bounding box with one additional parameter in an intuitive way. The 2D bounding box is the rectangle which encloses the 3D bounding box. 

The additional parameter denoted as $c_c$ is determined by measuring the vertical distance from the top of the 2D bounding box to the top frontal edge of the 3D bounding box and dividing it by the height of the 2D bounding box. This parameter thus always falls into the $\interval{0}{1}$ interval. The construction of the 2D bounding box and the additional parameter can be seen in Fig. \ref{fig:bb_construction}.

\begin{figure*}[]
   \centering
   \includegraphics[width=\textwidth]{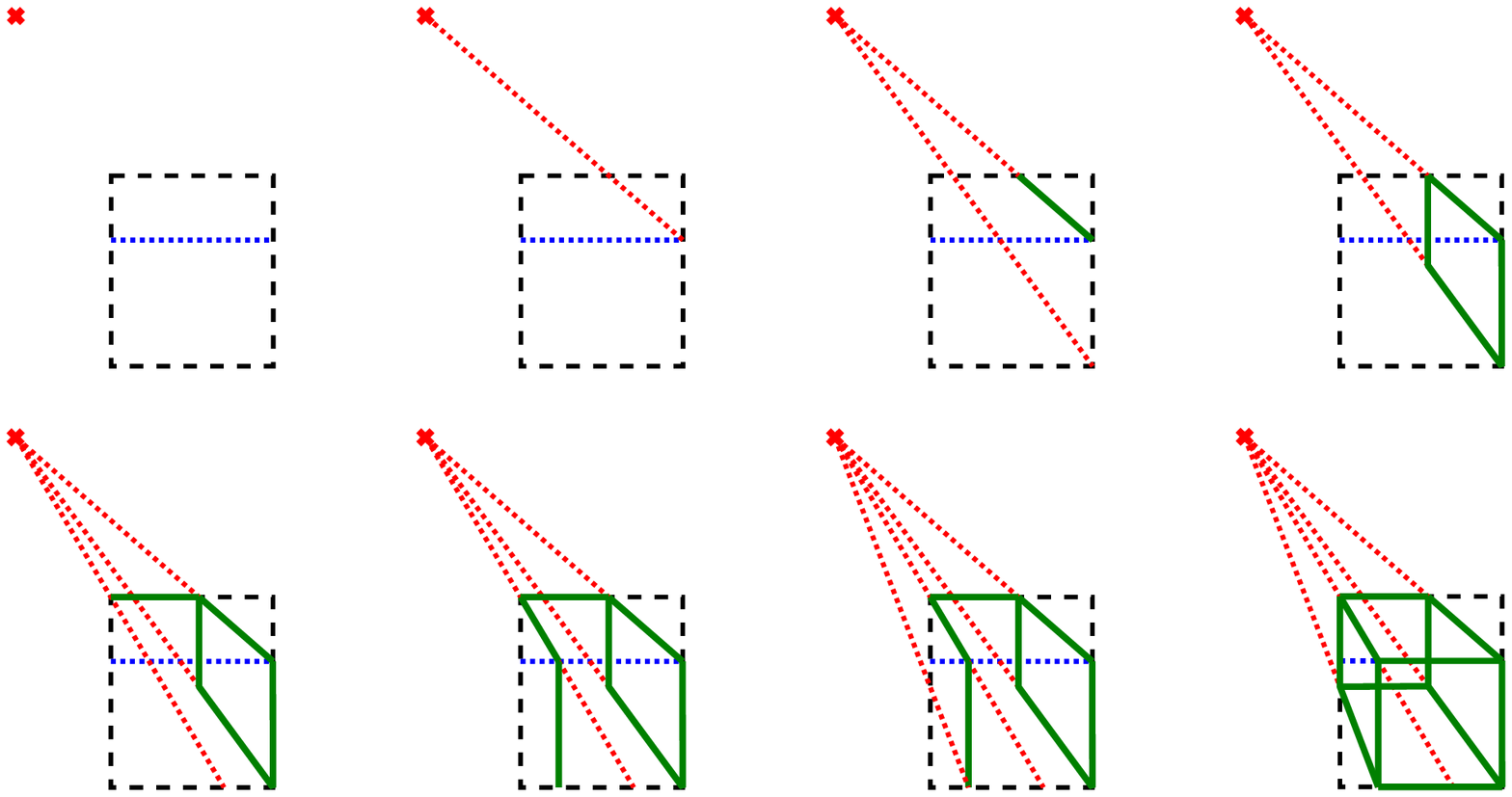}
   \caption{The process of reconstructing the 3D bounding box (solid green) from the known 2D bounding box (dashed black), the line given by the $c_c$ parameter (dotted blue) and the position of the \textit{VPU} (red cross). The process begins in the top left of the figure. Line segments originating in the \textit{VPU} (red dotted) are used when needed to determine the corners and edges of the 3D bounding box.}
   \label{fig:bb_reconstruction}
\end{figure*}

\subsection{3D Bounding Box Reconstruction}
\label{sec:VPU}

The 2D bounding with the $c_c$ parameter can be used to reconstruct the 3D bounding box. Here we apply a similar process to the one described in our previous work \cite{CVWW2019}, but we generalize it to accommodate to new cases which emerge from the improved way the perspective transformation is obtained. 

The process of reconstruction depends on the position of the vanishing point which was not used for the transformation. Note that the position of this vanishing point has to be known in the transformed image which is easily obtainable by applying the perspective transformation. For simplicity, we will denote this vanishing point in the transformed image as \textit{VPU}. 

Due to the geometry of the vanishing points there are only two possibilities of relative vertical positions of the 2D bounding box and \textit{VPU}. The box is either above or below the \textit{VPU}. Let us first consider that \textit{VPU} is above the 2D bounding box. 

In that case there are only three possibilities regarding the horizontal positions of \textit{VPU} and the 2D bounding box. If \textit{VPU} is to the right of the 2D bounding box then the left end of the line segment representing the $c_c$ is a vertex of the 3D bounding box. Knowing this vertex one can construct the 3D bounding box in the transformed image. Similarly, if the \textit{VPU} is to the left of the 2D bounding box then the right end of the line segment is used. If the \textit{VPU} is neither to the left or to the right of the 2D bounding box then either of them can be used as a corner to start construction of the 3D bounding box. The process is visualized in Fig. \ref{fig:bb_reconstruction} for the case when \textit{VPU} is to the left of the 2D bounding box.

In the case when the \textit{VPU} is below the 2D bounding box the process is almost identical, the only difference is that when \textit{VPU} is to the left of the 2D bounding box the left end of the $c_c$ line segment is used as a starting vertex and vice versa.

This process may fail to produce a valid 3D bounding box. This can be easily detected during the reconstruction process as in that case a part of at least one of the edges of the 3D bounding box would lie outside of the area enclosed by the 2D bounding box. This failure indicates that there is no valid 3D bounding box for the given parametrization and perspective geometry. Such a situation may occur as it is impossible to guarantee that a neural network outputs only valid outputs. Since this occurs only rarely a simple solution of regarding these outputs as false positives works well enough in practice.

After the 3D bounding box is constructed in the transformed image an inverse perspective transformation can be applied to the vertices of the 3D bounding box to obtain the 3D bounding box in the original image.

\subsection{Bounding Box Detection}

As shown in the previous subsection we only need to detect 2D bounding boxes with the parameter $c_c$. For this purpose we utilize the RetinaNet object detector \cite{RetinaNet}. This detector outputs 2D bounding boxes for the detected objects. We modify the method to add $c_c$ to each of the output boxes.

The RetinaNet \cite{RetinaNet}, as well as other object detecting meta-architectures, uses anchor boxes as default positions of bounding boxes to determine where the objects are. The object detection task is separated into three parts: determining which anchor boxes contain which objects, resizing and moving the anchor boxes to better fit the objects and finally performing non-maximum suppression to avoid multiple detections of the same object. To train the network a two-part loss (\ref{eqn:both_losses}) is used.

\begin{equation}
\label{eqn:both_losses}
L_{tot} = \frac{1}{N}\left(L_{conf} + L_{loc} \right)
\end{equation}

The loss is averaged over all $N$ anchor boxes, $L_{conf}$ is the Focal loss used to train a classifier to determine which objects, if any, are in the bounding box. $L_{loc}$ is the regression loss to train the network how to reshape and offset the anchor boxes. To include the parameter $c_c$ we simply add one additional regression loss which results in the total loss:

\begin{equation}
\label{eqn:all_losses}
L_{tot} = \frac{1}{N}\left(L_{conf} + L_{loc} + L_{c}\right).
\end{equation}

The loss $L_{c}$ (\ref{eqn:centers_loss}) is identical in the base structure to the loss used for the four regression parameters in the RetinaNet, which is itself based on the regression loss of the SSD object detector \cite{SSD}. The loss is calculated as a sum over all of the $N$ anchor boxes and $M$ ground truth bounding boxes. $x_{i,j}$ determines whether the $i$-th anchor box corresponds to the $j$-th ground truth label. We subtract the ground truth value of $c_c$ denoted as $c_{c,j}^g$ from the predicted value $c_{c,i}^p$ and apply the smooth L1 function ($s_{L1}$).

\begin{equation}
\label{eqn:centers_loss}
L_{c} = \frac{1}{N} \sum_{i=1}^N \sum_{j=1}^M x_{i,j} s_{L1} \left( c_{c,i}^p - c_{c,j}^g \right)
\end{equation}

Note that this approach could be extended to some of the more recent object detection frameworks which rely on keypoint detection \cite{CornerNet,ObjectsAsPoints}. However, we opt to use the anchor box based approach as this makes our method universally transferable between different object detection frameworks with widespread use. 

\subsection{Object Detector Training}

To obtain training data we use data from two distinct datasets. The first dataset is BoxCars116k \cite{boxcars}. The original purpose of this dataset is fine-grained vehicle classification. The dataset contains over 116 thousand images, each containing one car along with make and model labels, information on positions of vanishing points and the 3D bounding box of the car. We transform these images with the proposed transformation and calculate the 2D bounding boxes and the $c_c$ parameter based on the provided 3D bounding boxes. Since each image is only of one car we augment the images by randomly rescaling them and placing them on a black background.

\label{sec:BCS}

\begin{figure*}[t]
   \centering
   \includegraphics[width=1\textwidth]{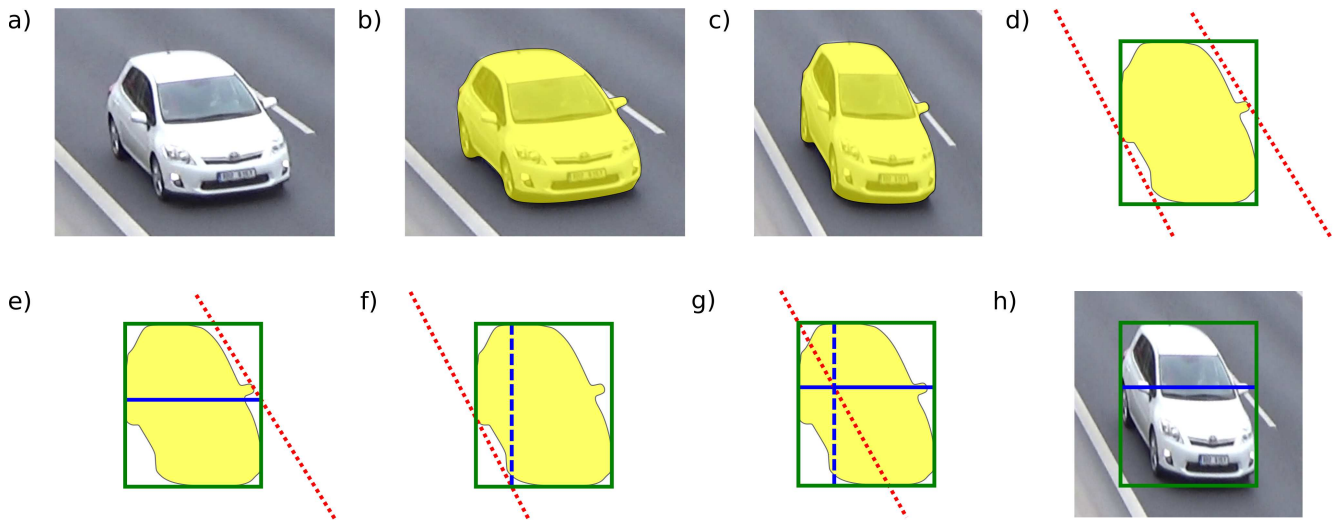}
   \caption{The process of creating annotations from provided mask of a vehicle. \textbf{a)} The original image. \textbf{b)} Mask (yellow) is obtained using Mask R-CNN \cite{MaskRCNN}. \textbf{c)} Both the mask and the image is transformed using the transformation for the pair \textit{VP2-VP3} (see subsection \ref{sec:VPU}). \textbf{d)} The 2D bounding box (green rectangle) and the two lines (dotted red) originating in \textit{VPU} and tangent to the mask are drawn. \textbf{e)} Intersection of one of the tangents with a vertical edge of the 2D bounding box can be used to determine the first candidate for the $c_c$ parameter line (solid blue). \textbf{f)} Intersection of the other tangent and the bottom edge of the bouding box is used to draw a vertical line (dashed blue) through it. \textbf{g)} A line (dotted red) originating in \textit{VPU} and going through the top-left corner of the bounding box is drawn. The intersection of this line and the vertical line from the previous step is used to determine the position of the $c_c$ parameter line (solid blue). \textbf{h)} Finally, from the two possible $c_c$ parameter lines depicted in e) and g) we choose the one constructed in g) as it creates a wider 3D bounding box. The result of this process is a 2D bounding box (green rectangle) with the $c_c$ parameter line in the transformed image. } 
   \label{fig:mask}
\end{figure*}

The other used dataset is BrnoCompSpeed \cite{brnocompspeed}. We use the split C of this dataset providing 9 videos for testing and 12 for validation and training. Each video is approximately one hour long with 50 frames per second (with one exception). For training and validation we use only every 25-th frame of the videos. We use the first 30000 frames for validation and the rest (140000-180000 depending on video length) are used for training. The main purpose of this dataset is to evaluate camera calibration and speed measurement algorithms. The cameras have been manually calibrated and thus the positions of the vanishing points are available, however the dataset does not contain 3D bounding box annotations.

To obtain the necessary 3D bounding box annotation we run these frames through Mask R-CNN \cite{MaskRCNN} image segmentation network trained on the COCO dataset \cite{COCO}. We transform the masks of detected vehicles and the images using our transformation and create the 2D bounding boxes with $c_c$ as labels for training. Obtaining a 2D bounding box from a mask is straightforward. The computation of the $c_c$ parameter requires a few steps since the masks may not be perfect and the cars are not commonly box shaped. The process begins with drawing the two lines tangent to the mask from both sides originating in \textit{VPU} (see subsection \ref{sec:VPU}). Each of these lines intersect the edges of the 2D bounding box twice. The intersection closer to the \textit{VPU} is discarded. Thus we have two points on the edges of the 2D bounding box each corresponding to one tangent line. Calculating the $c_c$ parameter for the point which lies on one of the vertical edges is straightforward. In case of a point on one of the horizontal edges of the bounding box a vertical line through this point is drawn. Next a line from \textit{VPU} is drawn to the closest corner of the 2D bounding box. The vertical position of this intersection is then used to determine the $c_c$ parameter. In the end we obtain two values for the $c_c$ parameter and use the one which creates a line closer to the \textit{VPU} thus choosing the wider of the two options. For visual reference see fig. \ref{fig:mask}.

\label{sec:variants}
Based on the development of the validation loss during the training we employ early stopping and train our models for 30 epochs each with 10000 training steps. For each pair of vanishing points we train models of three different sizes dependent on the input size of the transformed image. For the pair \textit{VP2-VP3} the sizes of the input image in pixels (width $\times$ height) are $960 \times 540, 640 \times 360$ and $480 \times 270$. For the pair \textit{VP1-VP2} we use the same dimensions we just flip them so the bigger dimension is the height of the image. We use the minibatch size of 16 for the models of the two smaller sizes and due to memory constraints a minibatch size of 8 for the largest model.

\subsection{Tracking}

From the object detector we obtain 2D bounding boxes with the parameter $c_c$ for vehicles in each frame of the recording. The tracking algorithm begins in the first frame with no active tracks and continues iterating through frames. For each 2D bounding box detected in the frame its IoU against the last 2D bounding box in each active track is calculated. If IoU of a detection is higher than 0.1 for at least one track, then the bounding box is added to the track with highest IoU score. If no track has at least 0.1 IoU against the detection, then a new active track is created. If a track hasn't had any bounding boxes added to it in the last 10 frames, then the track is no longer considered active and is added to the results. To detect speed we filter out bounding boxes which are less than 10 pixels away from the edges of the images. We also discard tracks which have less than 5 detected bounding boxes within them or smaller distance traveled than 100 pixels.

\subsection{Speed Measurement}

In the previous step the 2D bounding boxes with the parameter $c_c$ were grouped and filtered into relevant tracks. 3D bounding boxes are reconstructed (see subsection \ref{sec:VPU}) for all detections. Knowing the 3D bounding box position in the original image the speed is determined using a point which is in the middle of the frontal bottom edge of the 3D bounding box (see Fig. \ref{fig:bbox_qualitative}). Since these points should under normal circumstances lie on the road plane, we can use the camera calibration to easily determine the distances between various positions within a track. To detect the average speeds of the vehicles we employ the same method as \cite{brnocompspeed} by calculating the median of the interframe speeds of the whole track.

\section{Evaluation}

\begin{figure*}[]
   \centering
   \includegraphics[width=1\textwidth]{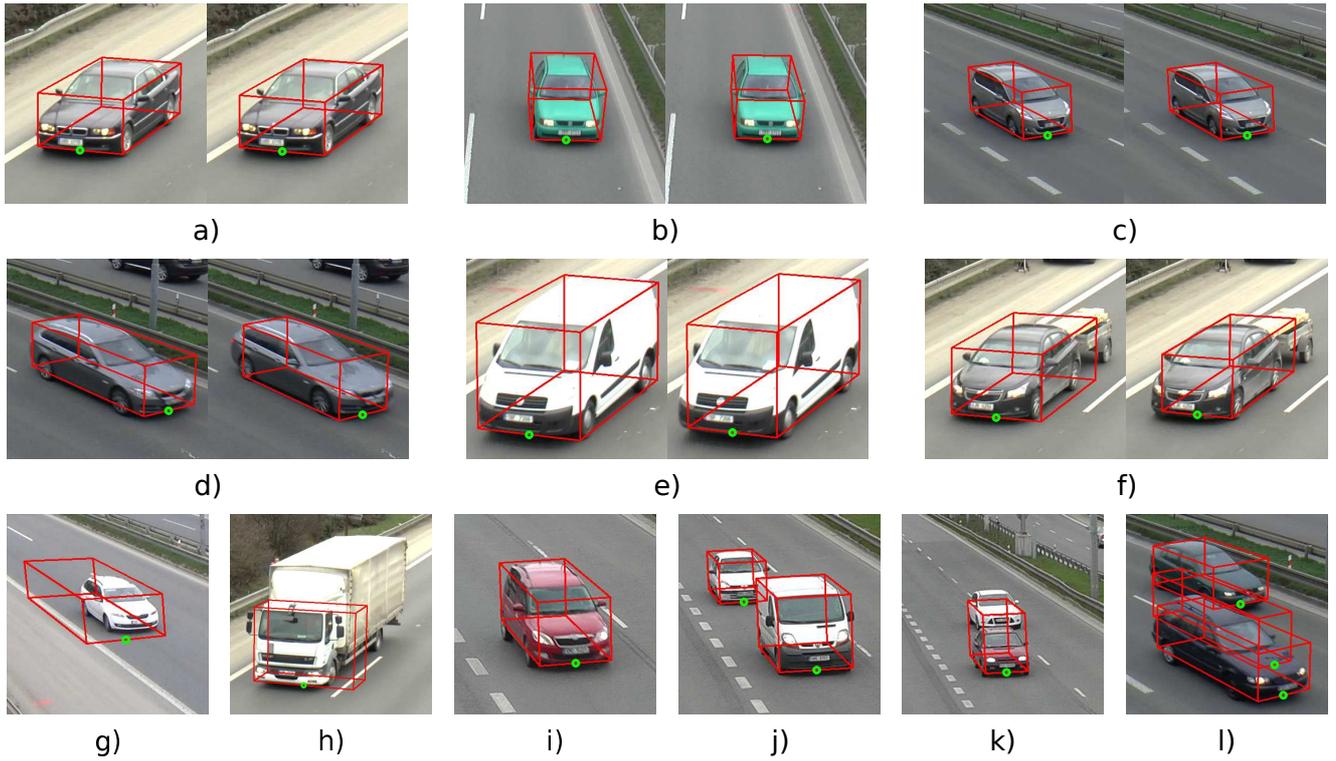}
   \caption{3D bounding boxes detected on the videos from the test set of the BrnoCompSpeed \cite{brnocompspeed} dataset. \textbf{a-f)} Results for the model using the pair \textit{VP2-VP3} and input size $640\times 360$ px are on the left and results for \textit{VP1-VP2} and $360 \times 640$ px are on the right. In images \textbf{a-c)} we can observe that there are only minor differences between the models and these predictions can be considered accurate, however \textbf{d-f)} show less accurate results. Inaccuracies for the pair \textit{VP1-VP2} are greater. \textbf{g-l)} Results for the pair \textit{VP2-VP3}. Similar results can be obtained also for the other pair. \textbf{g-i)} Various levels of inaccurate placement of the bounding boxes. \textbf{j)} An accurately detected occluded vehicle. \textbf{k)} Sometimes the two vehicles get grouped into one bounding box. \textbf{l)} Occlusion can also sometimes result in a false positive between the two vehicles, however such false positive usually gets filtered out during tracking.}
   \label{fig:bbox_qualitative}
\end{figure*}

\begin{table*}[]
\renewcommand{\arraystretch}{1.3}
\center
\caption{The results of the compared methods on the split C of the BrnoCompSpeed dataset \cite{brnocompspeed}.  Mean, median and 95-th percentile errors are calculated as means of the corresponding error statistics for each video. Recall and precision are averaged over the videos in the test set. FPS values for our methods are calculated on a machine with 8-core AMD Ryzen 7 2700 CPU, 48 GB RAM and Nvidia TITAN V GPU.}
\label{tab:speed_acc}
\resizebox{\textwidth}{!}{
\begin{tabular}{|c|c|c|c|c|c|c|c|c|}
\hline
Method                                & VP pair                           & \begin{tabular}[c]{@{}c@{}}Input Size\\ (px)\end{tabular} & \begin{tabular}[c]{@{}c@{}}Mean error\\ (km/h)\end{tabular} & \begin{tabular}[c]{@{}c@{}}Median error\\ (km/h)\end{tabular} & \begin{tabular}[c]{@{}c@{}}95-th percentile\\ (km/h)\end{tabular} & \begin{tabular}[c]{@{}c@{}}Mean precision\\ (\%)\end{tabular} & \begin{tabular}[c]{@{}c@{}}Mean recall\\ (\%)\end{tabular} & FPS         \\ \hline
\textit{DubskaAuto}\cite{dubska2014}    & \textit{-}                        & -                           & 8.22  & 7.87  & 10.43 & 73.48 & \textbf{90.08}   & -           \\ \hline
\textit{SochorAuto}\cite{sochor2017}    & \textit{-}                        & -                           & 1.10  & 0.97  & 2.22  & \textbf{90.72} & 83.34    & -           \\ \hline
\textit{SochorManual}\cite{sochor2017}  & \textit{-}                        & -                           & 1.04  & 0.83  & 2.35  & \textbf{90.72} & 83.34    & -           \\ \hline
\textit{Previous2D}\cite{CVWW2019}      & \multirow{2}{*}{\textit{VP2-VP3}} & \multirow{2}{*}{640 x 360}  & 0.83  & 0.60  & 2.17  & 83.53 & 82.06 & 62           \\ \cline{1-1} \cline{4-9} 
\textit{Previous3D}\cite{CVWW2019}      &                                   &                             & 0.86  & 0.65  & 2.17  & 87.67 & 89.32 & 62           \\ \hline
\multirow{6}{*}{\textit{Transform3D}}   & \multirow{3}{*}{\textit{VP2-VP3}} & 480 x 270                   & 0.92  & 0.72  & 2.35  & 89.26 & 79.99 & \textbf{70} \\ \cline{3-9} 
                                        &                                   & 640 x 360                   & 0.79  & 0.60  & 1.96  & 87.08 & 83.32 & 62          \\ \cline{3-9} 
                                        &                                   & 960 x 540                   & \textbf{0.75} & \textbf{0.58} & \textbf{1.84} & 87.74 & 83.21 & 43          \\ \cline{2-9} 
                                        & \multirow{3}{*}{\textit{VP1-VP2}} & 270 x 480                   & 1.12  & 0.84  & 2.84  & 87.68 & 84.06 & \textbf{70} \\ \cline{3-9} 
                                        &                                   & 360 x 640                   & 1.17  & 0.87  & 2.88  & 88.32 & 86.32 & 62          \\ \cline{3-9} 
                                        &                                   & 540 x 960                   & 1.09  & 0.84  & 2.65  & 88.06 & 85.30 & 43          \\ \hline
\multirow{2}{*}{\textit{Transform2D}}   & \textit{VP2-VP3}                  & 640 x 360                   & 0.92  & 0.69  & 2.18  & 84.73 & 77.58 & 62          \\ \cline{2-9} 
                                        & \textit{VP1-VP2}                  & 360 x 640                   & 1.11  & 0.91  & 2.70  & 86.96 & 79.42 & 62          \\ \hline
\textit{Orig2D}                         & -                                 & 640 x 360                   & 0.93  & 0.73  & 2.53  & 88.47 & 85.20 & 62          \\ \hline
\textit{MaskRCNN3D}                     & -                                 & 1024 x 576                  & 0.88  & 0.64  & 2.19  & 88.44 & 81.89 & 5          \\ \hline
\end{tabular}}
\end{table*}

The output 3D bounding boxes for two of our models are showcased in Fig. \ref{fig:bbox_qualitative} along with some cases where our models fail to detect the vehicle accurately. 

We also provide two videos showcasing our 3D detector on the first five minutes of the video titled \textit{session6\_center} from the test set of the split C of the BrnoCompSpeed dataset \cite{brnocompspeed}. In the first video (Online Resource 1) we used the vanishing point pair \textit{VP2-VP3} and the input size of $640 \times 360$ px. In the second video (Online Resource 2) we used the pair \textit{VP1-VP2} and the input size of $360 \times 640$ px. Note that in both of these videos we keep all of the detections to showcase some of the false positives, which are later removed during tracking.

\subsection{Speed Measurement Accuracy}

We evaluate our method on the speed measurement task on the split C of the BrnoCompSpeed dataset \cite{brnocompspeed}. The evaluation metrics can be seen in Table \ref{tab:speed_acc} and we provide files for evaluation for all of our presented variants including the ablation experiments online.\footnote{\url{https://github.com/kocurvik/BCS_results}} We compare our results to available published results on the same data. We include the original method by Dubsk\'{a} et al. \cite{dubska2014} denoted as \textit{DubskaAuto}. We also include its improved version by Sochor et al. \cite{sochor2017} in two variants: \textit{SochorAuto}, which to our knowledge is the most accurate fully automatic method for speed measurement evaluated on the dataset and \textit{SochorManual} which is more accurate, but includes a manual adjustment of the scale factor during calibration. We also include our previous work \cite{CVWW2019} in two variants: \textit{Previous3D}, which detects 3D bounding boxes aligned with the vanishing points and \textit{Previous2D}, which detects 2D bounding boxes aligned with just two of the vanishing points. Both of these methods require a manual adjustment of the perspective image transformation for some camera angles to work properly, therefore they can not be considered fully automatic. To our knowledge the results for the method denoted as \textit{Previous2D} are the best published so far with respect to the speed measurement accuracy on the dataset.

We report results for our method denoted as \textit{Transform3D} in its six variants described in subsection \ref{sec:variants}. For all of these we report the rate of frames per second that can be processed on a machine with 8-core AMD Ryzen 7 2700 CPU, 48 GB RAM and Nvidia TITAN V GPU. The results show that the variants with the two bigger input sizes using the pair of vanishing points $\textit{VP2-VP3}$ outperform all other published methods. The variants using the other pair of vanishing points show worse performance, but are still comparable to the results of \textit{SochorAuto}.

\subsection{Ablation Studies}

To properly gauge the impact of the perspective transformation we perform two ablation experiments. We train the standard RetinaNet 2D object detector on the same data as the other models, except that the images are not transformed. We refer to this model as \textit{Orig2D}. We also train the standard RetinaNet 2D object detector on the transformed images. We use the same 2D bounding boxes as in \textit{Transform3D}, but without the parameter $c_c$. We refer to this method as \textit{Transform2D}. We use the center of the bottom edge of the 2D bounidng box to determine the speeds. We train these models with the same hyperparameters as our base model. We perform the ablation experiments only for the image size of $640 \times 360$ and $360 \times 640$ pixels. We also perform an experiment denoted as \textit{MaskRCNN3D} where we obtain the 3D bounding boxes of vehicles via their mask obtained by using the Mask R-CNN network \cite{MaskRCNN} pre-trained on the MS COCO dataset \cite{COCO}. We construct the 3D bounding boxes in the same manner as described in subsection \ref{sec:variants}. The results of the ablation experiments can be seen in Table \ref{tab:speed_acc}.

When compared to the results \textit{SochorAuto} and \textit{SochorManual} which rely on Faster R-CNN in combination with background subtraction for detection it is clear that the use of RetinaNet alone (\textit{Orig2D}) for detection of vehicles brings significant improvements. Transforming the image (\textit{Transform2D}) also brings a minor improvement for the pair \textit{VP2-VP3}. It is clear that introducing the construction of the 3D bounding box for this pair improves speed measurement accuracy significantly and is thus beneficial for speed measurement tasks.

Surprisingly, the transformation for the pair \textit{VP1-VP2} increases the mean speed measurement error over the non-transformed variant. This may possibly be caused by the rectification of the image resulting in loss of some visual cues important for object localization. 

Results for \textit{MaskRCNN3D} are worse than the results of our main approach (\textit{Transform3D}) for the pair \textit{VP2-VP3}, but better than the results for other methods and ablation experiments. This ablation experiment provides further evidence that constructing a 3D bounding box is beneficial for the task of speed measurement, since it performed better than all of the three ablation setups which only used 2D bounding boxes. This ablation experiment required no task-specific training, however this came at a significant hit to the FPS performance and may thus not be a cost-effective option for real-world applications.

\subsection{Evaluating the influence of recall on the speed measurement accuracy}

\begin{table}[]
\renewcommand{\arraystretch}{1.3}
\center
\caption{The results of the compared methods on the subset of the split C of the BrnoCompSpeed dataset \cite{brnocompspeed} which contains only those ground truth vehicle tracks that were correctly detected by all of the compared methods. The original test set contains 13 704 tracks while the subset contains only 7 274 tracks. We present these results to indicate that the results in Table \ref{tab:speed_acc} are not skewed by difficult examples that are not detected by some of the methods with lower recall.}
\label{tab:speed_acc_full_recall}
\resizebox{0.49\textwidth}{!}{
\begin{tabular}{|c|c|c|c|c|}
\hline
Method                                & VP pair                           & \begin{tabular}[c]{@{}c@{}}Input Size\\ (px)\end{tabular} & \begin{tabular}[c]{@{}c@{}}Mean error\\ (km/h)\end{tabular} & \begin{tabular}[c]{@{}c@{}}Median error\\ (km/h)\end{tabular}          \\ \hline
\textit{DubskaAuto}\cite{dubska2014}    & \textit{-}                        & -                           & 8.16  & 8.35  \\ \hline
\textit{SochorAuto}\cite{sochor2017}    & \textit{-}                        & -                           & 1.05  & 0.90  \\ \hline
\textit{SochorManual}\cite{sochor2017}  & \textit{-}                        & -                           & 1.08  & 0.89  \\ \hline
\textit{Previous2D}\cite{CVWW2019}      & \multirow{2}{*}{\textit{VP2-VP3}} & \multirow{2}{*}{640 x 360}  & 0.71  & 0.56  \\ \cline{1-1} \cline{4-5} 
\textit{Previous3D}\cite{CVWW2019}      &                                   &                             & 0.72  & 0.58  \\ \hline
\multirow{6}{*}{\textit{Transform3D}}   & \multirow{3}{*}{\textit{VP2-VP3}} & 480 x 270                   & 0.84  & 0.69  \\ \cline{3-5} 
                                        &                                   & 640 x 360                   & \textbf{0.65}  & \textbf{0.53}  \\ \cline{3-5} 
                                        &                                   & 960 x 540                   & 0.68  & 0.55  \\ \cline{2-5} 
                                        & \multirow{3}{*}{\textit{VP1-VP2}} & 270 x 480                   & 1.16  & 0.87  \\ \cline{3-5} 
                                        &                                   & 360 x 640                   & 1.07  & 0.84  \\ \cline{3-5} 
                                        &                                   & 540 x 960                   & 1.05  & 0.90  \\ \hline
\multirow{2}{*}{\textit{Transform2D}}   & \textit{VP2-VP3}                  & 640 x 360                   & 0.72  & 0.59  \\ \cline{2-5} 
                                        & \textit{VP1-VP2}                  & 360 x 640                   & 1.05  & 0.87  \\ \hline
\textit{Orig2D}                         & -                                 & 640 x 360                   & 0.73  & 0.62  \\ \hline
\textit{MaskRCNN3D}                     & -                                 & 1024 x 576                  & 0.76  & 0.61  \\ \hline
\end{tabular}}
\end{table}

The speed measurement accuracy metrics presented in Table \ref{tab:speed_acc} could potentially be skewed due to variance in recall of the compared methods. A method with higher recall might be able to correctly detect more difficult instances of vehicles. These difficult examples could then lead to a higher speed measurement error which would not be the case for the methods with lower recall. To verify that this effect is not significant we tested all of the compared methods only on a subset of ground truth vehicle tracks which contained only those tracks that were correctly detected by all of the compared methods. This is equivalent to the largest subset of the ground truth tracks such that all of the compared methods would achieve 100 \% recall on it. This subset contains only 7 274 tracks compared to the standard amount of 13 704 tracks in the split C of the BrnoCompSpeed dataset \cite{brnocompspeed}.

We present the results evaluated on the subset in Table \ref{tab:speed_acc_full_recall}. It is clear that the subset indeed contains easier examples as the accuracy of almost all of the compared methods on the subset is slightly better than on the original test set. The relative performance of the compared methods remains similar to the results on the full test set. Our method \textit{Transform3D} for the pair \textit{VP2-VP3} still outperforms the other compared methods including the results for the ablation experiments. Interestingly, the variant which uses the input size of $640 \times 360$ px outperforms the variant with the input size of $960 \times 640$ px. The recall of these two variants on the full test set is very similar so we assume that the variant with smaller input size can perform better on easier examples, but worse in more difficult cases thus leading to its worse accuracy on the full test set.

\subsection{Computational Costs}

All of our variants run at faster than real-time (25 FPS) speeds, though we have to note that the testing videos of the BrnoCompSpeed dataset were recorded with 50 FPS and the speed measurement accuracy can therefore be worse for footage with lower FPS rates. Results show that increasing the input image size tends to result in increased speed measurement accuracy, while also increasing the computational demands reflected by the FPS rates for models of different sizes. Our method is therefore easily configurable to work under different hardware constraints in real-world applications. We were not able to perform FPS measurement for the other published approaches, however the most significant methods \textit{SochorAuto} and \textit{SochorManual} both rely on the Faster R-CNN object detector which, in general, is significantly slower than RetinaNet used in our method \cite{RetinaNet}.

\subsection{Evaluation on dataset by Luviz\'{o}n et al.}

To verify that our method is capable of working in various traffic surveillance scenarios we performed an evaluation on the dataset published along with \cite{luvizon}. In comparison with the BrnoCompSpeed dataset \cite{brnocompspeed} there is only one intersection filmed under various weather conditions. The ground truth speeds are measured using inductive loops located in the top portion of the scene and the camera is positioned closer to the road plane.

Since this dataset contains only vehicles moving away from the camera, whereas our models were mostly trained on vehicles moving towards the camera, we fine-tuned the \textit{Transform3D} model for the \textit{VP2-VP3} pair and the input size of $960 \times 540$ on training data obtained from this dataset in the same way we produced training data for the BrnoCompSpeed dataset (see subsection \ref{sec:BCS}) and the BoxCars116k dataset \cite{boxcars} for 10 epochs. The authors of the dataset provide no official split for training and testing so we perform tests on the first half of videos in each subset\footnote{The dataset contains five subsets of videos. For the subsets which contain an odd number of videos we use the odd video for testing.} and use the remaining videos for training.

The speed measurement method proposed in \cite{luvizon} uses a separate camera calibration for each of the three surveilled lanes. To make a reasonable comparison we use only one set of vanishing points, but we use a different scale for each of the lanes. Our method achieves recall of 98.9\% for the vehicles with valid ground truth measurements with 92.7\% of measured speeds falling within the range of -3 to +2 km/h speed measurement error proposed as the evaluation metric for the dataset. In comparison, the method proposed in \cite{luvizon} achieves better results with 99.2\% and 96\% respectively on the whole dataset. We consider our results to be competitive since our method has fewer limitations such as not requiring the camera to be so close to the road plane for the license plates to be clearly visible and aligned with the camera as well as not requiring a manual calibration.

\section{Conclusion}

We proposed several improvements and extensions to our previously published method \cite{CVWW2019} for detection of 3D bounding boxes of vehicles in a traffic surveillance scenario. Our improvements eliminate the need to manually adjust the construction of the perspective transformation for some camera angles. We also extended the transformation method to enable using a different pair of vanishing points. 

We have also extended the experimental analysis of our method providing a range of configurations, which allows for a flexibility of choice with respect to the accuracy-speed tradeoff for real-world applications. All of the models can be run in real-time on commercially available GPUs. Configurations relying on smaller input sizes provide a possibility of processing multiple video streams concurrently.

Our improved fully automatic approach led to an improvement in speed measurement on the BrnoCompSpeed dataset \cite{brnocompspeed}. Compared to our previously published non-automatic state of the art method \cite{CVWW2019} we reduced the mean speed measurement error by 10\% (0.83 km/h to 0.75 km/h), the median speed measurement error by 3\% (0.60 km/h to 0.58 km/h) and the the 95-th percentile error by 15\% (2.17 km/h to 1.84 km/h). Compared to the state of the art fully automatic method \cite{sochor2017} we reduced the mean absolute speed measurement error by 32\% (1.10 km/h to 0.75 km/h), the absolute median error by 40\% (0.97 km/h to 0.58 km/h) and the 95-th percentile error by 17\% (2.22 km/h to 1.84 km/h). 

\begin{acknowledgements}
The authors would like to thank Adam Herout for his valuable comments.  The authors also gratefully acknowledge the support of NVIDIA Corporation with the donation of GPUs.
\end{acknowledgements}


\bibliographystyle{spmpsci}      % mathematics and physical sciences
\bibliography{main}   % name your BibTeX data base


\end{document}